\documentclass[letterpaper, 10 pt, conference]{ieeeconf}  

\IEEEoverridecommandlockouts                              

\usepackage{graphics} 
\usepackage{epsfig} 
\usepackage{mathptmx} 
\usepackage{times} 
\usepackage{amsmath} 
\usepackage{amssymb}  
\usepackage{graphicx}
\usepackage{multicol}
\usepackage{hyperref}
\usepackage{algorithm}
\usepackage{algpseudocode}
\usepackage{amsmath}
\usepackage{booktabs}
\usepackage{multirow}
\usepackage{makecell}
\usepackage{subcaption}
\usepackage{tabularx}
\usepackage{float}

\title{\LARGE \bf
OCEAN: An Openspace Collision-free Trajectory Planner for Autonomous Parking Based on ADMM 

}

\author{Dongxu Wang$^\dag$$^{1}$, Yanbin Lu$^\dag$$^{1}$, Weilong Liu$^{1}$, Hao Zuo$^{1}$, Jiade Xin$^{1}$, Xiang Long$^{1}$, Yuncheng Jiang$^{2}$ 
\thanks{\dag Authors contribute equally to this paper.}
\thanks{$^{1}$D. Wang, Y. Lu, W. Liu, H. Zuo, and J. Xin are with Mach Drive, LLC, Shanghai, China (Corresponding author: H. Zuo
{\tt\small hao.zuo@mach-drive.com}).}%
\thanks{$^{2}$ Y. Jiang is with the Department of Electronic and Computer Engineering, Hong Kong University of Science and Technology, Hong Kong(email: {\tt\small yjiangbw@connect.ust.hk}). }%
}

\begin{document}

\maketitle
\thispagestyle{empty}
\pagestyle{empty}

\begin{abstract}
In this paper, we propose an $\underline{O}$penspace $\underline{C}$ollision-fre$\underline{E}$ tr$\underline{A}$jectory pla$\underline{N}$ner (OCEAN) for autonomous parking. OCEAN is an optimization-based trajectory planner accelerated by Alternating Direction Method of Multiplier (ADMM) with enhanced computational efficiency and robustness, and is suitable for all scenes with few dynamic obstacles. Starting from a hierarchical optimization-based collision avoidance framework, the trajectory planning problem is first warm-started by a collision-free Hybrid A* trajectory, then the collision avoidance trajectory planning problem is reformulated as a smooth and convex dual form, and solved by ADMM in parallel. The optimization variables are carefully split into several groups so that ADMM sub-problems are formulated as Quadratic Programming (QP), Sequential Quadratic Programming (SQP), and Second Order Cone Programming (SOCP) problems that can be efficiently and robustly solved. We validate our method both in hundreds of simulation scenarios and hundreds of hours of public parking areas. The results show that the proposed method has better system performance compared with other benchmarks.

\end{abstract}

\section{INTRODUCTION}
Autonomous driving has long been a hot topic, and autonomous parking systems are drawing rising interest among consumers. More recently, Automated Valet Parking (AVP) has attracted increasing attention both from industry and academia. It enjoys higher level of autonomy, and can drive a car from the entry of the parking lot to the parking spot autonomously. 

However, AVP also brings more challenges to autonomous parking system. Firstly, classical curve-based parking planner \cite{curve_parking} can no long deal with cluttered and dense parking environment during parking spot searching. Secondly, due to non-linear and non-holonomic vehicle dynamics and non-convexity of the free space, planning a collision-free trajectory is nontrivial. Thirdly, the computation time grows rapidly with the number of obstacles, and it makes most accurate collision evaluation algorithms hard to satisfy real-time requirement in practice.  

Special attention is given to collision evaluation in autonomous parking. To simplify collision avoidance evaluation, one way is to consider the point-mass object model \cite{point_mass_collision_avoidance}, but such simplification involves large approximation errors. Safe corridors, which has been widely used in unmanned aerial vehicles, can generate collision-free space consisting of a series of connected polygons(or polyhedrons). However, such expression inevitably sacrifices available driving areas, and becomes too conservative in dense and cluttered parking environments.

To tackle accurate collision evaluation in parking planning, obstacles and ego vehicle should be represented in full dimension, and the collision avoidance problem is formulated as a Mixed-Integer Programming (MIP) \cite{MIP}. Despite various approaches to solve MIP problem, most of they cannot satisfy simultaneous requirement. Therefore, \cite{linear_prog} proposed a whole-body evaluation which is formulated as a low-dimensional linear programming. However, this method is only applicable to convex obstacles and cannot selectively compute distances to different obstacles. Euclidean Signed Distance Field (ESDF) is also widely used in collision evaluation \cite{esdf}, but practical applications often require a trade-off between map resolution and computational time, and ESDF is seldom used in autonomous parking due to its high memory and computation resource consumption.

Hybrid Optimization-based Collision Avoidance (H-OBCA) which reformulates collision avoidance constraints as its dual form by introducing additional dual variables is proposed in \cite{h_obca} and \cite{obca}. Although this approach converts the original non-convex and nonlinear optimization problem to a convex and nonlinear one, it is unacceptable to make the dimension of the problem increase dramatically. Recent improvements based on OBCA \cite{obca} is dual warm starts with Reformulated OBCA (TDR-OBCA), and have shown promising results in real world implementation \cite{tdr}. However, the computational time is still far from satisfactory. To further accelerate OBCA, \cite{rda} proposed Regularized Dual Alternating Direction Method of Multiplier (RDA) to decompose the large centralized problem into several sub-problems by adopting ADMM. These sub-problems can be solved in a distributed optimization framework. However, RDA only split optimization variables with dual variables, thus having limited improvement in computational efficiency.

To sum up, the challenges of autonomous parking planning comes from 1) accurate collision avoidance evaluation, and 2) requirement of algorithmic real-time performance and robustness. Inspired by OBCA, we present an robust and accelerated collision-free motion planner named as OCEAN to deal with prescribed challenges. This method aims to accurately evaluate collision risk, and has better real-time performance and robustness than other benchmark methods. The main contribution of this paper is as follows:

\begin{itemize}
\item Our method is based on Model Prediction Control (MPC) framework to embed non-holonomic vehicle dynamics and control limits in problem formulation. By leveraging dual convex reformulation, the non-convex collision-avoidance constraints are reformulated as convex constraints.

\item ADMM is adopted to decompose the original nonlinear programming problem into a series of modified QP, SQP, and SOCP problems. These problems can be solved in fully decentralized and parallel architecture so that computation burden is greatly alleviated and real-time performance can be assured.

\item Unlike H-OBCA, TDR-OBCA and RDA, which rely heavily on warm start results, our method reformulates the dual variable collision avoidance problem as a SOCP problem, and iteratively solve it in ADMM structure. Warm start is not necessary in our method.

\item To deal with non-linearity in state transition, we introduce additional optimization variables and constraints to linearize the nonlinear constraints in respective sub-problems.

\item  We validate the efficiency and robustness of the algorithm both in hundreds of simulation scenarios and hundreds of hours of public parking areas. As soon as this paper is published, the source code of the proposed parking algorithm will be released as soon as this paper is published.
\end{itemize}



This paper is organized as follows. Section \ref{sec:related work} reviews related work. Section \ref{sec:problem formulation} describes the trajectory optimization problem and its dual reformulation of collision-free constraints. The main algorithm of OCEAN is shown in Section \ref{sec:algo}. Section \ref{sec: simulation} shows the simulation and real-world test results, and conclusion is presented in Section \ref{sec:conclusion}.

\section{RELATED WORK}\label{sec:related work}

In the field of autonomous driving planning algorithms, the MPC based trajectory planning method has been widely promoted and applied \cite{mpc}. However, in multi-obstacle parking environments, the non-convex collision avoidance constraints pose a challenge to trajectory optimization algorithms. Currently, there are several obstacle modelling methods applicable to trajectory optimization:

\subsection{Ellipsoids or Sphere Approximation}
The first approach models obstacles as ellipsoids in 2D or spheres in 3D\cite{spheres}. This method is intuitive and simple, as it permits direct calculation of obstacle distances and determination of the gradient direction away from obstacles. However, it does not accurately represent the distance between the vehicle and the obstacles.

\subsection{ Safe Corridor}
The second approach utilizes the concept of safe corridors \cite{corridor}\cite{corridor2}, commonly employed in unmanned aerial vehicles, to redefine the obstacle space as a convex polygonal constraint space. This approach guarantees trajectory safety within the confines of the convex space. However, the cost associated with constructing these corridors depends on the resolution of the obstacle discretization and the step size used during corridor expansion. Consequently, these factors lead to conservative estimations of the actual obstacle space. Moreover, the construction of high-quality corridors becomes expensive when the vehicle trajectory is in close proximity to obstacles.


\subsection{Double Signed Distance Obstacle Representation}
The third approach of this study adopts the double signed distance obstacle representation and dual convex reformation.\cite{double sd}. In this approach, obstacles are modeled as convex polyhedral shapes, and the distance between the vehicle and the obstacles is represented by signed distance. By using this representation, the minimum distance $d_{\min}$ to each obstacle can be directly incorporated. Then by leveraging dual convex reformulation, the collision avoidance constraints are converted to nonlinear but convex constraints. The vehicle model can be either a point mass or a convex polyhedron. However, the double signed distance obstacle representation introduces a significant number of equality and inequality constraints to the problem, thereby increasing its dimension. For a point mass system scenario with $N$ sampling points and $M$ obstacles, it introduces $N^M$ dual variables. When the shape of the controlled system is considered, the dimension of the problem increases dramatically, resulting in an additional $2N^M$ linear inequality constraints.

\subsection{Linear Programming}
The fourth method formulates the optimization problem of the distance between the vehicle plane and the obstacle planes in the whole-body collision as a linear programming (LP) problem \cite{linear_prog}. This significantly improves the computational efficiency by using only one variable to ensure trajectory safety, regardless of the number of obstacles. However, this method is only applicable to convex obstacles and cannot selectively compute distances to different obstacles. In addition, it is mainly combined with trajectory optimisation algorithms based on polynomial representations.

\subsection{Euclidean Signed Distance Field}
The fifth method is based on potential field approaches, such as ESDF \cite{esdf} and risk fields \cite{risk field}. However, practical applications often require a trade-off between resolution and computational time. Recently, the SDF Any-Shape method \cite{sdf any} has been proposed to compute the signed distance of the scanning volume formed by the robot and its trajectory implicitly, in a delayed and efficient manner by exploiting the spatio-temporal continuity. This method is mainly combined with trajectory optimization problems based on polynomial representations. However, it requires the trajectory to be differentiable with respect to time.

\subsection{Accelerate Optimization Algorithms}
Existing approaches primarily focus on reducing computation time through algorithm design aspects. These approaches incorporate various techniques such as heuristics, approximations, parallel computation, learning, and edge computing \cite{edge computing}.
This paper aims to accelerate problem solving by effectively formulating warm-start problems and decomposing the centralized optimization problem through ADMM \cite{admm}.
It is noteworthy that ADMM, a straightforward yet potent algorithm, has been widely used in many fields \cite{admm}\cite{admm ilqr}.
By applying parallel computation to obstacle avoidance, ADMM has been shown to accelerate autonomous navigation in cluttered environments \cite{rda}. Since the collision avoidance constraints are non-convex, and ADMM can only be applied to a convex problem, we should make proper approximation and reformulation to convexify the original problem before applying ADMM.

\section{PROBLEM FORMULATION}\label{sec:problem formulation}

Inspired by OBCA, we formulate the parking planning problem in the MPC framework, and reformulate the collision-avoidance constraints as its dual form so that the original MIP problem is transformed to a MPC problem with nonlinear and convex constraints. 
\subsection{Vehicle Dynamic Model}
We use bicycle model to denote ego state. At time step $k$, the ego state is denoted by $\mathbf{x_k} = [X_k,Y_k,\varphi_k,v_k]\in \mathbb{R}^{4}$, where $X_k$ and $Y_k$ are vehicle lateral and longitudinal position, $\varphi_k$ is heading angle, and $v_k$ is vehicle velocity. The control command is $\mathbf{u_k} = [\delta_k, a_k]$, where $\delta_k$ is the steering and $a_k$ is the acceleration. the vehicle dynamics in discrete form is given by

\begin{equation} \label{equ_1}
\begin{aligned}
\begin{array}{ll} 
X_{k+1}=X_{k}+v_{k} t_{k} \cos \varphi_{k}, \\
Y_{k+1}=Y_{k}+v_{k} t_{k} \sin \varphi_{k}, \\
\varphi_{k+1}=\varphi_{k}+v_{k} t_{k}  \frac{\tan \delta_{k}}{L}, \\
v_{k+1}=v_{k}+a_{k} t_{k}, \\
\forall k=1, \ldots, N,                         
\end{array}
\end{aligned}
\end{equation} 

where $t_k$ is variable time step, and $L$ is the wheelbase.

\subsection{Full Dimension Modeling and Collision Avoidance}
For reader's convenience, we follow the same notation as that in OBCA. Readers can refer to \cite{h_obca}(Sec II) for more details. For a given state $x_k$, we denote by $\mathbb{E}(x_k) \subset \mathbb{R}^{2}$ the space occupied by ego vehicle, and $\mathbb{O}^{(m)}\subset \mathbb{R}^{2}, \forall m=1, \ldots, M$ the space occupied by obstacles. The ego vehicle initial rectangle ${\mathbb{B}}$ and the rotated and translated rectangle $\mathbb{E}(x_k)$ is defined as 
\begin{equation}
\begin{aligned}
\begin{array}{ll} 
\mathbb{E}(x_k)=R(x_k) \mathbb{B}+t(x_k), \quad \mathbb{B}:=\{y: G y \leq g\}.
\end{array}
\end{aligned}
\end{equation} 
where $G$ and $g$ define initial rectangle ${\mathbb{B}}$ and are assumed to be known. Similarly, obstacles are defined in the same way
\begin{equation}
\begin{aligned}
\begin{array}{ll} 
\mathbb{O}^{(m)}=\left\{y \in \mathbb{R}^{2}: A^{(m)} y \leq b^{(m)}\right\}, \forall m=1, \ldots, M
\end{array}
\end{aligned}
\end{equation} 

The collision free constraints can be defined as $\mathbb{E}(x) \cap \mathbb{O}^{(m)}=\emptyset, \forall m=1, \ldots, M$, which is nonlinear and non-convex. We use signed distance \cite{double sd} to analytically formulate collision avoidance
\begin{equation}
\begin{aligned}
\begin{array}{ll} 
\operatorname{sd}(\mathbb{E}(x), \mathbb{O}):=\operatorname{dist}(\mathbb{E}(x), \mathbb{O})-\operatorname{pen}(\mathbb{E}(x), \mathbb{O}),
\end{array}
\end{aligned}
\end{equation} 
where the distance and penetration functions are respectively defined as
\begin{subequations}
\begin{equation}
\begin{aligned}
\begin{array}{ll} 
\operatorname{dist}(\mathbb{E}(x), \mathbb{O}):=\min _{t}\{\|t\|:(\mathbb{E}(x)+t) \cap \mathbb{O} \neq \emptyset\}, \\
\end{array}
\end{aligned}
\end{equation}
\begin{equation}
\begin{aligned}
\begin{array}{ll} 
\operatorname{pen}(\mathbb{E}(x), \mathbb{O}):=\min _{t}\{\|t\|:(\mathbb{E}(x)+t) \cap \mathbb{O}=\emptyset\}.
\end{array}
\end{aligned}
\end{equation}
\end{subequations}

According to \cite{obca}, the collision avoidance constraints can be converted to its dual form by introducing additional dual variables which is reformulated as follows:
\begin{equation} \label{dual_form}
\begin{aligned}
\begin{array}{ll} 
\operatorname{sd}(\mathbb{E}(x), \mathbb{O})>d \\
\Longleftrightarrow \quad \exists \lambda \geq 0, \mu \geq 0:-g^{\top} \mu+(A t(x)-b)^{\top} \lambda>d, \\
\quad G^{\top} \mu+R(x)^{\top} A^{\top} \lambda=0,\left\|A^{\top} \lambda\right\|=1 .
\end{array}
\end{aligned}
\end{equation} 
 Then we modify the non-convex constraint \eqref{dual_form} as a convex one for simplification. Since we have robust warm-started trajectory that is collision free, such modification is reasonable:
\begin{equation} \label{dual_form_simplification}
\begin{aligned}
\begin{array}{ll} 
\operatorname{sd}(\mathbb{E}(x), \mathbb{O})>d \\
\Longleftrightarrow \quad \exists \lambda \geq 0, \mu \geq 0:-g^{\top} \mu+(A t(x)-b)^{\top} \lambda>d, \\
\quad G^{\top} \mu+R(x)^{\top} A^{\top} \lambda=0,\left\|A^{\top} \lambda\right\| \le 1 .
\end{array}
\end{aligned}
\end{equation}

\subsection{Problem Statement}
We formulate the autonomous parking problem in MPC framework, and the state evolution in discrete form is as follows
\begin{equation} \label{state_trans}
\begin{aligned}
\begin{array}{ll} 
\mathbf{x}_{k+1}=\mathbf{A}_{t} \mathbf{x}_{k}+\mathbf{B}_{k} \mathbf{u}_{k}+\mathbf{c}_{k}, 
 \quad \forall k=1, \cdots, N,
\end{array}
\end{aligned}
\end{equation} 
where $\mathbf{A}_{t}$ and $\mathbf{B}_{k}$ are coefficient matrices that can be easily obtained from \eqref{equ_1}. We also enforce constraints on control $\mathbf{u}_{\min } \preceq \mathbf{u}_{k} \preceq \mathbf{u}_{\max }$, and higher order control can also be applied limit by finite discretization $\mathbf{\hat{u}_{\min }} \preceq \mathbf{\hat{u}_{k}} \preceq \mathbf{\hat{u}_{\max}}$, where $\hat{u} = \frac{\mathbf{u}_{k+1}-\mathbf{u}_{k}}{t_k}$. State constraints can also be applied when necessary. We combine these constraints as
\begin{equation} \label{cost_ieq}
\begin{aligned}
\begin{array}{ll} 
h(x_k,u_k,u_{k-1}) \le 0, \forall k=1, \cdots, N.
\end{array}
\end{aligned}
\end{equation} 

The goal of the optimization problem is to plan a smooth and safe parking trajectory. Therefore, we define objective function as follows
\begin{equation}\label{cost_obj}
\begin{aligned}
\begin{array}{ll} 
\displaystyle \mathcal{L}(x_k, u_k, t_k) &= w_{x} \sum\limits_{k=1}^{N} \|x_k-\tilde{x}_k\|_{2}^{2} \\
& +w_{u}\sum\limits^N_{k=1}\|u_k\|_{2}^{2} \\
& +w_{x^{\prime}} \sum\limits^N_{k=1}\frac{\|x_k-x_{k-1}\|_{2}^{2} }{t_k} \\
& +w_{\bar{u}}\sum\limits^N_{k=1}\frac{\|u_k-\hat{u}_{k-1}\|_{2}^{2}}{t_k},
\end{array}
\end{aligned}
\end{equation} 
where $\tilde{x}$ represents reference state from warm start, and $\hat{u}$ represents the previous step control commands. In \eqref{cost_obj}, the first term penalizes state difference w.r.t warm start trajectory. The second term penalizes control efforts. The third term measures second order smoothness, and the forth term penalizes state difference between consecutive optimization loops. 

To this end, the parking planning can be formulated as an optimization problem with objective function \eqref{cost_obj}, subjecting to constraints \eqref{dual_form_simplification}\eqref{state_trans}\eqref{cost_ieq}. However, such formulation only set lower bound for collision avoidance, the parking trajectory is supposed to be away from obstacles as far as possible. Therefore, we introduce $d$ in \eqref{dual_form_simplification} as a slackness variable, and plugin it into objective function in a exponent form $\mathcal{M}\left(d\right) = e^d$. Now we can give the exact form of our optimization problem

\begin{subequations}\label{optimzation_prob}

\begin{equation}\label{exact_obj}
\begin{aligned}
\min_{x,u,d,t,\lambda,\mu}\;\; & \sum^N_{k=1}\mathcal{L}(x_k,u_k,t_k)+ w_d\sum_{k=1}^{N}\sum_{m=1}^M\mathcal{M}(d_{k,m}),  
\end{aligned}
\end{equation} 
\begin{equation}\label{cons_bound}
\begin{aligned}
s.t.  \;\;&x_0=x_S,x_{N+1}=x_F,  \\  
\end{aligned}
\end{equation} 
\begin{equation}\label{cons_state_trans}
x_{k+1}=f(x_k,u_k),  \\ 
\end{equation} 
\begin{equation}\label{cons_ieq}
h(x_k,u_{k-1})\le 0,  \\
\end{equation}
\begin{equation}\label{cons_d}
 d_{k,m} \le 0, \\
\end{equation}
\begin{equation}\label{cons_collision_1}
-g^T\mu_{k,m}+(A_m t_{k}-b_{m})^T\lambda_{k,m}+d_{k,m}=0,  \\
\end{equation}
\begin{equation}\label{cons_collision_2}
G^T\mu_{k,m}+R(x_k)^TA_{m}^T\lambda_{k,m}=0,  \\
\end{equation}
\begin{equation}\label{cons_lamda}
\left\|A_m^T\lambda_{k,m}\right\|_*\le 1,  \\
\end{equation}
\begin{equation}\label{cons_dual_variable}
\lambda_{k,m}\succeq_{k^*}0,\mu_{km}\succeq_{\bar{k}^*}0,  \\
\end{equation}
\end{subequations}
\begin{equation*}
\begin{aligned}
 \forall k=1,...,N,m=1,...,M,  \\
\end{aligned}
\end{equation*} 
where $w_d$ is penalty weight for $d$, $x_S$ the initial state, and $x_F$ the final state.

\section{ COLLISION-FREE TRAJECTORY PLANNER}\label{sec:algo}
In this section, we introduce our parallel computation framework. To accelerate our trajectory planning problem, which introduced a large number of dual variables, we carefully split the original problem into several sub-problems so that each problem can be solved in parallel. And each sub-problem is well organized as a QP, or SOCP, or SQP problem which can be solved efficiently and robustly.


\subsection{Parallel Computation} \label{parallel_computation}
We use ADMM to solve \eqref{optimzation_prob}, where each iteration includes solving a series convex sub-problems. In each iteration, both dual variables and decision variables are carefully split into several sub-problems so that the problem can be solved efficiently and the variables can be updated in parallel.

We first give the augmented Lagrangian of \eqref{optimzation_prob} as follows
\begin{equation} \label{lagrangian}
\begin{aligned}
\mathcal{L}_\rho(x, u,t,d,&\lambda,\mu)=\sum^N_{k=1}\mathcal{L}(x_k,u_k,t_k)+\sum_{k=1}^{N}\sum_{m=1}^M\mathcal{L}(d_{k,m})  \\
&+\sum_{k=1}^{N}I_m(x_k,u_k,t_k) \\
&+\sum_{k=1}^{N}\sum_{m=1}^MI_{k,m}(\lambda_{km},\mu_{km},d_{k,m})  \\
&+\frac{\rho}{2}\sum_{k=1}^{N}\left\| f(x_k,u_k)-x_{k+1}+\eta_k \right\|_2^2  \\
&+\frac{\rho}{2}\sum_{k=1}^{N+1}\sum_{m=1}^M\rVert G_{k,m}(x_k,\lambda_{k,m},\mu_{k,m},d_{k,m})+\zeta_{k,m}\lVert_2^2  \\
&+\frac{\rho}{2}\sum_{k=1}^{N+1}\sum_{m=1}^M\rVert H_{k,m}(x_k,\lambda_{k,m},\mu_{k,m})+\xi_{k,m}\lVert_2^2  \\
\end{aligned}
\end{equation} 
where ${\eta_k,\zeta_{k,m},\xi_{k,m}}$ are dual variables corresponding to the equality constraints \eqref{cons_state_trans} \eqref{cons_collision_1} and \eqref{cons_collision_2}, and $\rho$ is the penalty parameter of Lagrangian function. We rewrite \eqref{cons_bound} \eqref{cons_ieq} as $I_m(x_k,u_k,t_k)$ and \eqref{cons_lamda} \eqref{cons_dual_variable} as $I_{k,m}(\lambda_{km},\mu_{km},d_{k,m})$  where  $I(.)$ is indicator function, i.e. $I(.)=0$ if corresponding constraints are satisfied and $I(.)= \infty$, otherwise. $G_{k,m}(.)$ and $H_{k,m}$(.) are simplified form of \eqref{cons_collision_1} and \eqref{cons_collision_2}
\begin{equation} \label{}
\begin{aligned}
\begin{array}{ll}
G_{k,m}(x_k,\lambda_{k,m},\mu_{k,m},d_{k,m})&=d_{k,m} -g^T\mu_{k,m} 
 \\
& +(At(x_k)-b_{m})^T\lambda_{k,m}, 
\end{array}
\end{aligned}
\end{equation} 
\begin{equation} \label{}
\begin{aligned}
\begin{array}{ll}
H_{k,m}(x_k,\lambda_{k,m},\mu_{k,m})=G^T\mu_{k,m}+R(x_k)^TA_{m}^T\lambda_{k,m}.
\end{array}
\end{aligned}
\end{equation} 

In \cite{rda}, ADMM is used to solve a similar parking planning problem, however, the variables are only decomposed into two groups, and the parallel computation is not efficient. We decompose the variables into four groups 
 $[\{\lambda, \mu\},\{v, a\},\{t\},\{x, y, \varphi, \delta, d, \sin \varphi, \cos \varphi, \tan \delta\}]$, 
 so that each sub-problem can be solved efficiently and robustly. Note that the dual variables $\{\lambda, \mu\}$ are only related to collision avoidance constraints, and according to \eqref{cons_collision_1}, the slackness variables $d$ are closely related to dual variables. Therefore, we first split the dual variables and slackness variables. For simplicity, we omit the subscript $(.)_{k,m}$, and we introduce superscript $(.^i)$  and $(.^{i+1})$ to indicate current step and next step variables respectively.
\begin{equation} \label{admm_1}
\begin{aligned}
\begin{array}{l} 
\{\lambda^{i+1},\mu^{i+1},d^{i+1}\}=\displaystyle \arg\min_{\lambda,\mu,d} \;\; \mathcal{L}_\rho(\{ \lambda,\mu \},\{ v^i,a^i \}, \{ t^i \},\\
 \{x^i,y^i,\varphi^i,\delta^i,d^i,\sin\varphi^i,\cos\varphi^i,\tan\delta^i \}).
\end{array}
\end{aligned}
\end{equation} 
To solve \eqref{admm_1}, one can use SQP by warm-starting the dual variables $\mu_{k,m}$ and $\lambda_{k,m}$. It is shown in \eqref{dual_form_simplification} that the dual variables are only related to collision avoidance constraints, i.e. \eqref{cons_collision_1}-\eqref{cons_dual_variable}, therefore, we can calculate $\mu_{k,m}$ and $\lambda_{k,m}$ by introducing a slackness variable $d$ to define the warm start problem with collision avoidance constraints
\begin{equation} \label{cost_ieq_1}
\begin{aligned}
\begin{array}{l} 
\displaystyle \min_{\mu, \lambda, d}\;\;\sum_{m=1}^{M} \sum_{k=1}^{K} d_{m,k} \\
s.t. -g^{T} \mu_{m,k}+\left(A_{m} t_k-b_{m}\right)^{T} \lambda_{m,k}
+d_{m,k}=0, \\
 G^{T} \mu_{m,k}+R\left(\tilde{x}^{*}(k)\right)^{T} A_{m}^{T} \lambda_{m,k}=0, \\
 \lambda_{m,k}\succeq 0, \mu_{m,k} \succeq 0, d_{m,k}<0 ,
\left\|A_{m}^{T} \lambda_{m,k}\right\|_{2} \le 1\\
  \quad \forall  k=1, \ldots K, m=1, \ldots, M,
\end{array}
\end{aligned}
\end{equation} 
And \eqref{cost_ieq_1} is equivalent to \eqref{admm_1}. Note that \eqref{cons_lamda} is a second order cone constraint, and to simplify the warm start problem, \cite{tdr} put the constraint $ \left\|A_{m}^{T} \lambda_{m,k}\right\|_{2} \le 1$ into objective function. Then the warm start problem is given by
\begin{equation} \label{cost_ieq_2}
\begin{aligned}
\begin{array}{l} 
\displaystyle \min _{\boldsymbol{\mu}, \boldsymbol{\lambda}, \boldsymbol{d}} \;\; \frac{1}{\beta} \sum_{m=1}^{M}\left\|A_{m}^{T} \lambda_{m}(k)\right\|_{2}^{2}+\sum_{m=1}^{M} \sum_{k=1}^{K} d_{m,k}\\
s.t. \quad -g^{T} \mu_{m,k}+\left(A_{m} t_k-b_{m}\right)^{T} \lambda_{m,k}
+d_{m,k}=0, \\
G^{T} \mu_{m,k}+R\left(\tilde{x}^{*}(k)\right)^{T} A_{m}^{T} \lambda_{m,k}=0, \\
\lambda_{m,k} \succeq 0, \mu_{m,k} \succeq 0, d_{m,k}<0 ,\\
\quad \forall  k=1, \ldots K, m=1, \ldots, M,
\end{array}
\end{aligned}
\end{equation} 
where $\frac{1}{\beta} $ is penalty weight, and the original problem becomes a QP problem. Although it is efficient to warm start dual variables by formulating a QP problem, \eqref{cost_ieq_2} alters the original optimization problem \eqref{cost_ieq_1}. As a result, solving \eqref{cost_ieq_2} may not necessarily guarantee strict collision avoidance.

In OCEAN, we pose \eqref{admm_1} into SOCP, which can be solved by existing primal-dual interior point method (PMIPM) and does not depend on warm start. IPM algorithms for SOCP are guaranteed to converge to at least a local optima \cite{ipm}. In \eqref{admm_1}, the optimization variables $\{\lambda^{i+1},\mu^{i+1},d^{i+1}\} $ have explicit geometric interpretation, and by posing the problem as a SOCP, we can expect more robust and accurate solutions. According to \cite{app_socp}\cite{mars landing}, there are SOCP solvers with deterministic convergence properties which means problems can be solved in polynomial time, and the global optimum can be computed with a deterministic upper bound on the number of iterations. 


Vehicle dynamic constraints \eqref{cons_state_trans} are nonlinear, we split decision variables ${v, a}$ so that \eqref{cons_state_trans} is converted to linear constraints
\begin{equation} \label{admm_2}
\begin{aligned}
\left\{v^{i+1}, a^{i+1}\right\} &= \arg \min_{v, a} \; \mathcal{L}_{\rho}\left(\left\{\lambda^{i+1}, \mu^{i+1}\right\},\{v, a\},\left\{t^{i}\right\},\right. \\
& \left.\left\{x^{i}, y^{i}, \varphi^{i}, \delta^{i}, d^{i}, \sin \varphi^{i}, \cos \varphi^{i}, \tan \delta^{i}\right\}\right)
\end{aligned}
\end{equation} 
Similarly, the variable $t$ is only related to the nonlinear constraints \eqref{cons_state_trans} and \eqref{cons_collision_1}, we split $t$ so that \eqref{cons_state_trans} and \eqref{cons_collision_1} become linear
\begin{equation} \label{admm_3}
\begin{aligned}
\left\{t^{i+1}\right\} &= \arg \min_{t} \mathcal{L}_{\rho}\left(\left\{\lambda^{i+1}, \mu^{i+1}\right\},\left\{v^{i+1}, a^{i+1}\right\},\{t\},\right. \\
& \left.\left\{x^{i}, y^{i}, \varphi^{i}, \delta^{i}, d^{i}, \sin \varphi^{i}, \cos \varphi^{i}, \tan \delta^{i}\right\}\right)
\end{aligned}
\end{equation} 
The optimization problem decision variable are$\{x, y, \varphi, \delta, d \}$, and to linearize vehicle dynamic constraints \eqref{cons_state_trans}, we introduce $ \{\sin \varphi, \cos \varphi, \tan \delta \}$ and enforce additional constraints to ensure linearization. Meanwhile, in \eqref{cons_collision_2}, the rotation matrices $R(x_k)$ contain nonlinear terms, and by introducing $ \{\sin \varphi, \cos \varphi\}$, \eqref{cons_collision_2} is also converted to linear constraints. Let us define $\mathcal{X} := \{x, y, \varphi, \delta, d, \sin \varphi, \cos \varphi, \tan \delta\}$ and $\mathcal{X}$ is split as follows
\begin{equation} \label{admm_4}
\begin{aligned}
\mathcal{X}^{i+1} &= \arg \min_{\mathcal{X}} \mathcal{L}_{\rho}(\{\lambda^{i+1}, \mu^{i+1}\},\{v^{i+1}, a^{i+1}\},\{t^{i+1}\},\mathcal{X})
\end{aligned}
\end{equation}
Then additional slackness constraints are introduced to indicate correlation between $\{ \varphi,\delta \}$ and $\{\sin \varphi, \cos \varphi, \tan \delta \} $ by using first order Taylor expansion. We introduce superscript $(.^j)$ and $(.^{j+1})$ to indicate current step and next step variables respectively in the SQP iteration. Specifically, $\{\sin \varphi, \cos \varphi, \tan \delta \} $ are linearized around the respective variables from last optimization loop as follows
\begin{equation} \label{admm_4_1}
\begin{aligned}
\begin{array}{ll} 
\sin \varphi^{j+1}-\varphi^{j+1} \cos \varphi^{j}=\sin \varphi^{j}-\varphi^{j} \cos \varphi^{j} \\
\cos \varphi^{j+1}+\varphi^{j+1} \sin \varphi^{j}=\cos \varphi^{j}+\varphi^{j} \sin \varphi^{j} \\
\tan \delta^{j+1}-\delta^{j+1} \sec ^{2} \delta^{j}=\tan \delta^{j}-\delta^{j} \sec ^{2} \delta^{j}
\end{array}
\end{aligned}
\end{equation} 
Since the slackness variables $d$ cannot be fully split, as both \eqref{admm_1} and \eqref{admm_4} can update $d$, we will update $d$ in both for better convergence.

Lastly, ADMM dual variables  are updated as follows
\begin{equation} \label{admm_update}
\begin{aligned}
\begin{array}{ll} 
\eta_{k}^{i+1} & =\eta_{k}^{i}+f\left(x_{k}^{i+1}, u_{k}^{i+1}\right)-x_{k+1}^{i+1} \\
\xi_{k, m}^{i+1} & =\xi^{i}+H_{k, m}\left(x_{k}^{i+1}, \lambda_{k, m}^{i+1}, \mu_{k, m}^{i+1}\right), \\
\zeta_{k ,m}^{i+1} & =\zeta_{k ,m}^{i}+G_{k, m}\left(x_{k}^{i+1}, \lambda_{k ,m}^{i+1}, \mu_{k ,m}^{i+1}, d_{k, m}^{i+1}.\right)
\end{array}
\end{aligned}
\end{equation} 

The stopping criteria for terminating the iteration is given as follow
\begin{subequations}
\begin{align}
\label{admm_stop_1}
&\sum_{k=1}^{N}\left\| f(x_k,u_k)-x_{k+1} \right\|_2^2  
+ \sum_{k=1}^{N}\sum_{m=1}^M \left\| G_{k,m}(x_k,\lambda_{k,m},\mu_{k,m},d_{k,m})\right\|_{2}^{2} \nonumber \\
&+ \sum_{k=1}^{N}\sum_{m=1}^M \left\|H_{k,m}(x_k,\lambda_{k,m},\mu_{k,m})\right\|_{2}^{2} \leq \epsilon^{\mathrm{pri}}, \\
\label{admm_stop_2}
&\begin{aligned}
&\sum_{k=1}^{N}\left\|\eta_{k}^{i+1}-\eta_{k}^{i}\right\|_{2}^{2} \\
&+ \sum_{k=1}^{N}\sum_{m=1}^M\left[\left\|\xi_{k, m}^{i+1}-\xi_{k, m}^{i}\right\|_{2}^{2}
+\left\|\zeta_{k, m}^{i+1}-\zeta_{k, m}^{i}\right\|_{2}^{2}\right] \leq \epsilon^{\text {dual }}.
\end{aligned}
\end{align}
\end{subequations}
where \eqref{admm_stop_1} guarantee the prime residual being small, and \eqref{admm_stop_2} guarantee the dual residual being small. The algorithm is given in Alg. \ref{alg1}.

\begin{algorithm}[t]
    \caption{OCEAN Planner Algorithm}
    \label{alg1}
    
    \begin{algorithmic}[1]
        \State 
            Warm start state and control variables by hybrid A* trajectory
         \For{$ iteration\quad k =1,2,\dots,$}
        \State  Update variables $\{\lambda^{i+1},\mu^{i+1},d^{i+1}\}$ by solving \eqref{admm_1}
       \State  Update variables $\left\{v^{i+1}, a^{i+1}\right\}$ by solving \eqref{admm_2}
       \State  Update variables $\left\{t^{i+1}\right\}$ by solving \eqref{admm_3}
        \State Update variables $\left\{\mathcal{X}^{i+1}\right\}$ by solving  \eqref{admm_4} with addition constraints \eqref{admm_4_1}
       \State Update Lagrangian multipliers ${\eta_k,\zeta_{k,m},\xi_{k,m}}$ by \eqref{admm_update}
        \If{\eqref{admm_stop_1} and \eqref{admm_stop_2} are satisfied} 
        \State break
        \EndIf
       
        \EndFor
      
    \end{algorithmic}
        
\end{algorithm}

\section{SIMULATION AND RESULTS}\label{sec: simulation}
In this section, we validate OCEAN in both simulation and real world road tests. We first conduct comparison between OCEAN and benchmark methods, i.e. H-OBCA and TDR-OBCA in \ref{simulation_2} under the same scenario setting and initial states, and we will show how OCEAN works efficiently and robustly. In \ref{simulation_3}, we simulate OCEAN under different scenarios and initial states, and in \ref{simulation_4} we show real world test results of OCEAN. The code is developed using C++ and deployed on an Intel(R) Xeon(R) Gold 6130 CPU processor with a clock speed of 2.10GHz. 

\begin{figure}[ht]
    \centering

    \begin{subfigure}[b]{0.22\textwidth}
        \includegraphics[width=\textwidth]{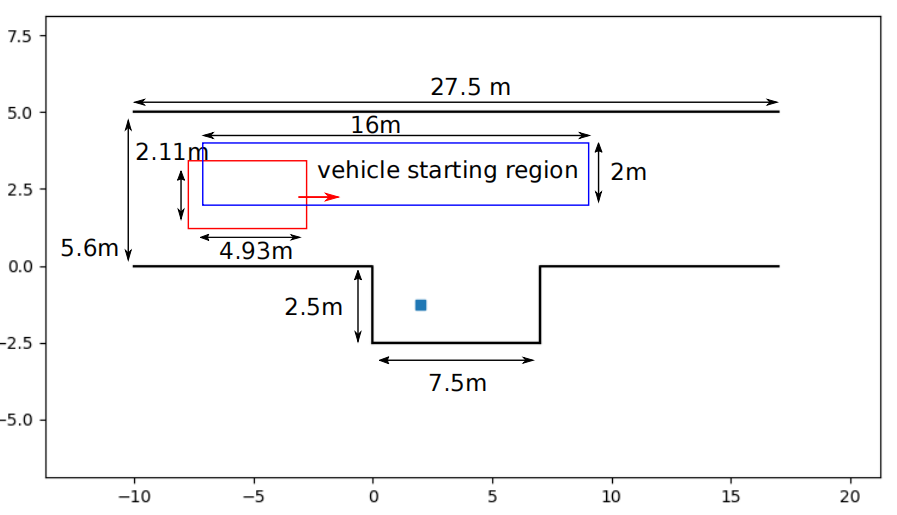}
        \caption{
		\label{fig:parallel_sim}
		Parallel Parking Scene
	}
    \end{subfigure}
    \hfill
    \begin{subfigure}[b]{0.22\textwidth}
        \includegraphics[width=\textwidth]{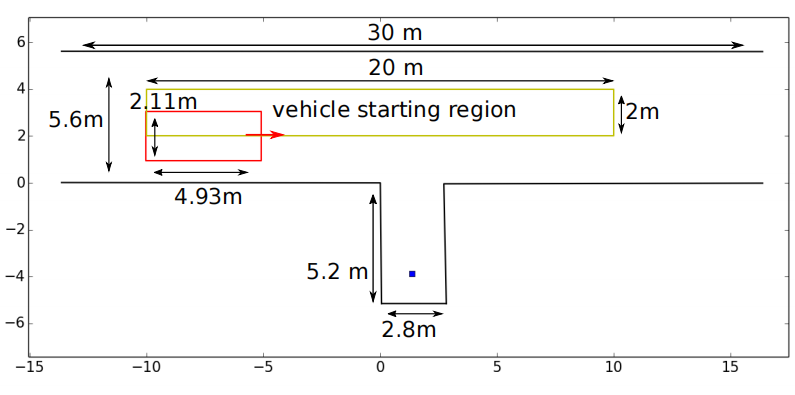}
       \caption{
		\label{fig:apollo_vertical}
		Vertical Parking Scene
	}
    \end{subfigure}

    \caption{Benchmark Scenarios.}
    \label{benchmark scenarios}
\end{figure}

\begin{figure}[ht]
    \centering

    \begin{subfigure}[b]{0.22\textwidth}
        \includegraphics[width=\textwidth]{{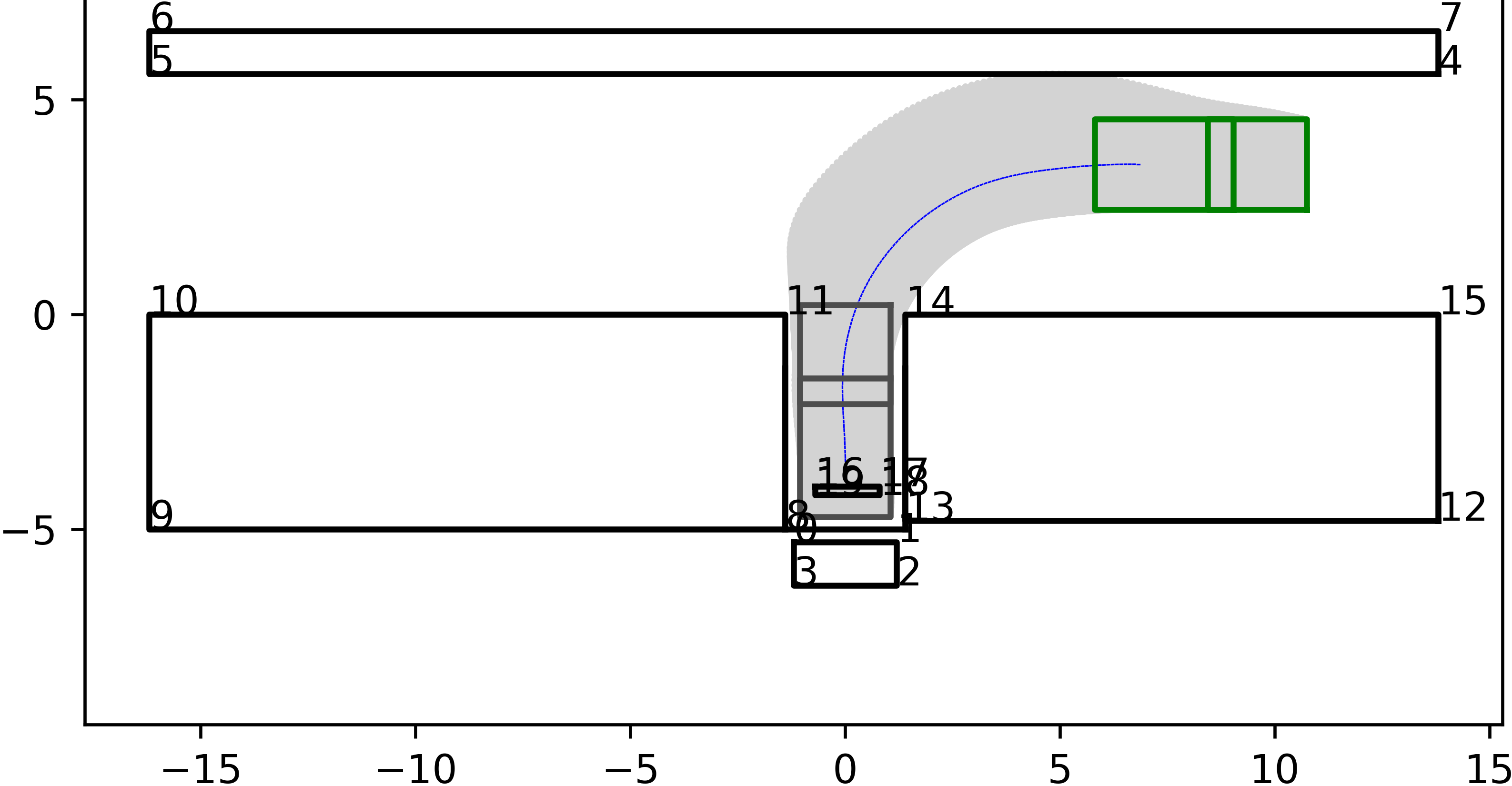}}
        \caption{H-OBCA}
    \end{subfigure}
    \hfill
    \begin{subfigure}[b]{0.22\textwidth}
        \includegraphics[width=\textwidth]{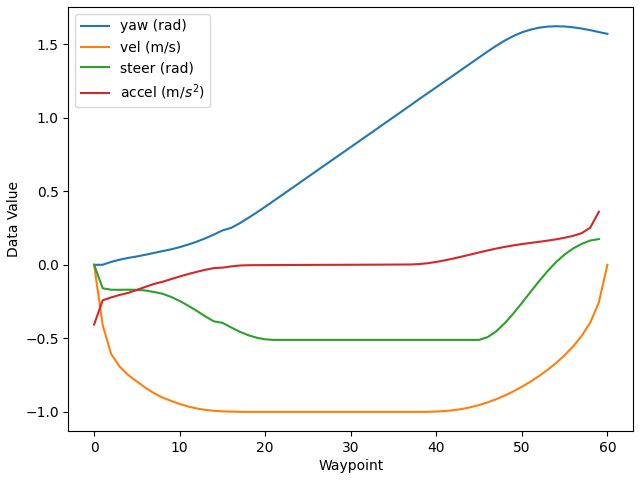}
        \caption{H-OBCA}
    \end{subfigure}
    
    \vspace{1em} 

    \begin{subfigure}[b]{0.22\textwidth}
        \includegraphics[width=\textwidth]{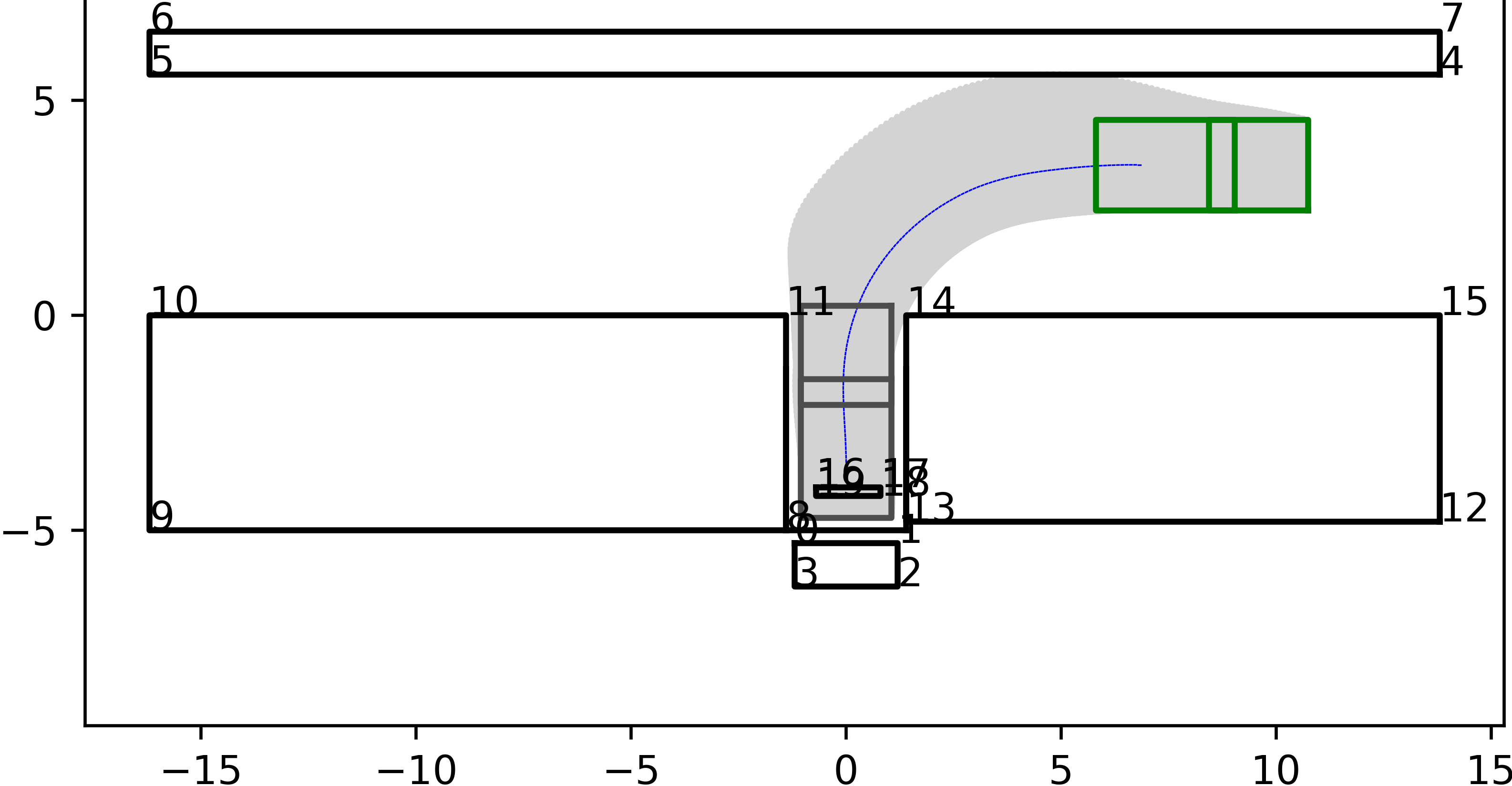}
        \caption{TDR-OBCA}
    \end{subfigure}
    \hfill
    \begin{subfigure}[b]{0.22\textwidth}
        \includegraphics[width=\textwidth]{{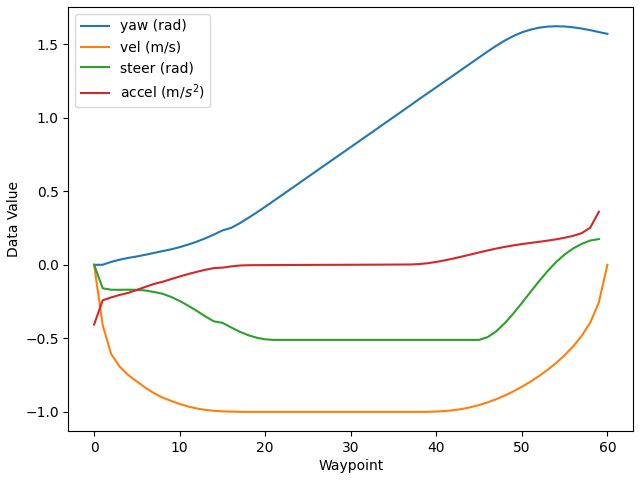}}
        \caption{TDR-OBCA}
    \end{subfigure}

    \vspace{1em} 

    \begin{subfigure}[b]{0.22\textwidth}
        \includegraphics[width=\textwidth]{{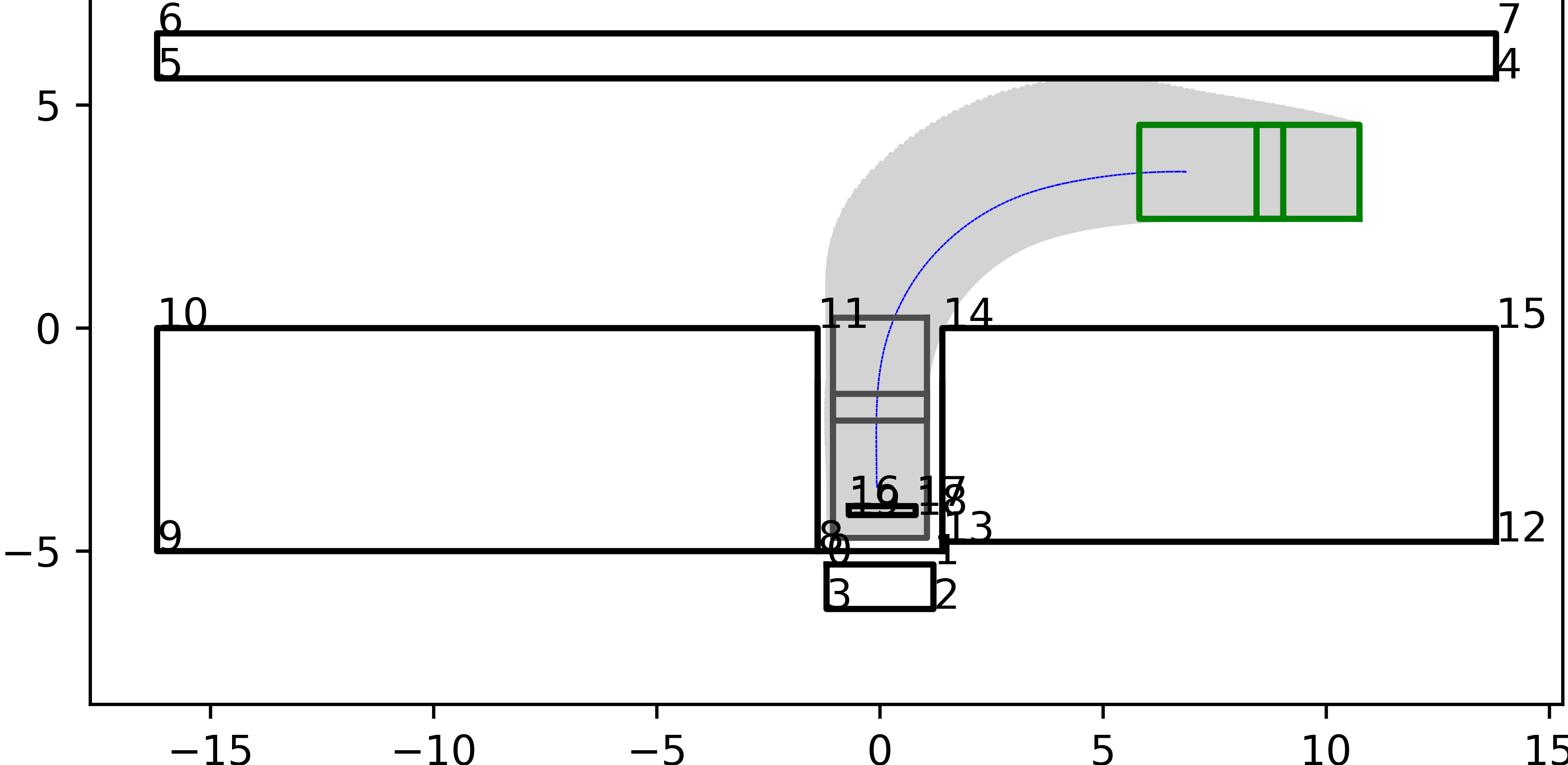}}
        \caption{OCEAN}
    \end{subfigure}
    \hfill
    \begin{subfigure}[b]{0.22\textwidth}
        \includegraphics[width=\textwidth]{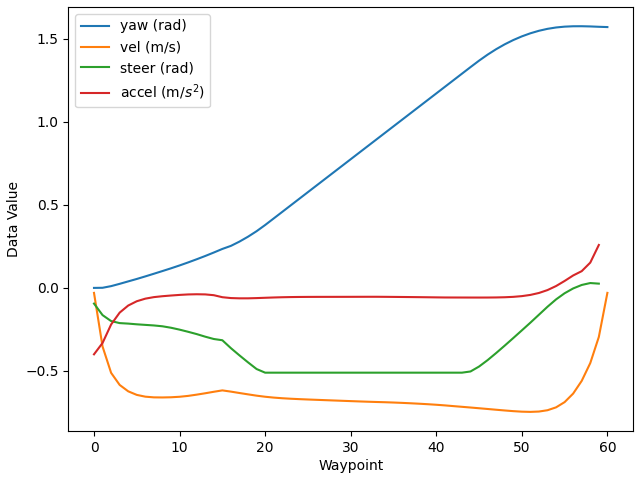}
        \caption{OCEAN}
    \end{subfigure}

    \caption{Parking Trajectory Visualization and Kinematic Parameters Sampling.}
    \label{fig:apollo_vertical_dyn}
\end{figure}

\begin{figure*}[]
	\centering
	\includegraphics[scale=0.18]{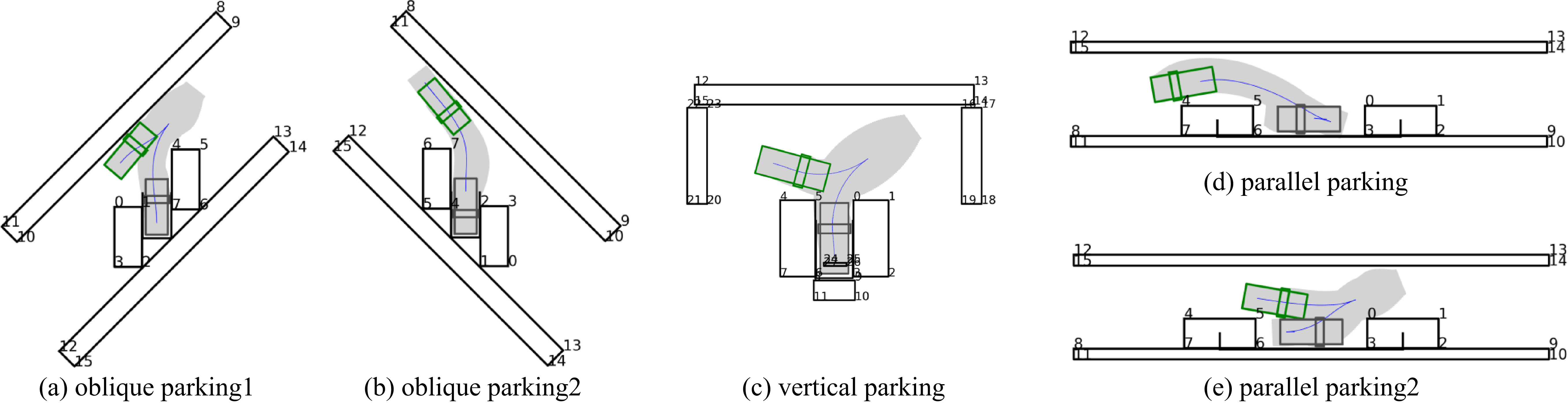}
	\vspace{0.1cm}
    \caption{Planning trajectories in simulation including scenarios: 1) oblique; 2) vertical; 3) parallel.}
    \label{fig:our_sim_test}
	\vspace{1em}
\end{figure*}
\subsection{Benchmark Simulation Results}\label{simulation_1}
For fair comparison, we try to restore the simulation results of other benchmark methods to the greatest extent. We set up a parallel parking scenario as shown in Fig.\ref{fig:parallel_sim} which is the same as that in \cite{iaps}, and performed tests for H-OBCA. The resulting system performance and time consumption are consistent with that in \cite{iaps}.

For TDR-OBCA, we use CasADi automatic differentiation and IPOPT (with Ma27 as the linear solver) to solve the NLP problem, and OSQP to warm start the dual problem. As a result, the warm start problem takes an average of 13ms to solve, and the NLP problem takes 1.39s to solve. Overall, TDR-OBCA achieves an 11.5\% efficiency improvement over H-OBCA in the parallel parking scenarios, and the time consumption of H-OBCA and TDR-OBCA are similar to that in \cite{tdr}\cite{iaps}.

\subsection{Comparison with Benchmark Methods}\label{simulation_2}
In order to compare OCEAN with H-OBCA and TDR-OBCA, we set up a vertical parking scenario as illustrated in Fig.\ref{fig:apollo_vertical}, which is the same as the one described in \cite{tdr}\cite{iaps}. Meanwhile, we use the same front-end rough path generation techniques for all methods so that the optimization results are not affected by initial guess. Besides, we adopt an improved hybrid A* method which generates path with continues curvature \cite{rs}.

The proposed method shows significant advantages in both control feasibility and computational efficiency. The simulation includes 80 scenarios defined by different initial positions of the ego vehicle \cite{tdr}. The initial positions are uniformly sampled from the yellow grid as shown in Fig.\ref{fig:apollo_vertical}. The initial heading angle is set to zero pointing to the right, and the final parking position remains constant for all scenarios. To demonstrate efficiency, we compared the running time of the three algorithms, and the results are shown in Table.\ref{tab:fail_rates}. 

\begin{table}
  \caption{Comparison of Three Algorithms: Failure Rate, Runtime Comparison, and Time Reduction Ratio Compared to H-OBCA.}
  \label{tab:fail_rates}
  \centering
  \begin{tabular}{cccc}
    \toprule
    Algorithm & Failure Rate & Runtime(ms) & Reduced Rate \\
    \midrule
    H-OBCA   & 0\% & 2723 & \ \\
    TDR-OBCA & 0\% & 2431 & 11.5\% \\
    OCEAN    & 0\%    & 233 & 91\% \\
    \bottomrule
  \end{tabular}
\end{table}

According to the experiments, we find that as long as hybrid A* search provides collision-free initial results, both H-OBCA and TDR algorithms generally do not fail during the optimization phase. The main difference lies in the 11.5\% reduction in solving time achieved by using warm-start technique to solve the dual problem.


\begin{table}
    \centering
    \captionsetup[table]{skip=20pt} 
    \caption{Trajectory evaluation results over the successful cases.}
    \label{tab:compare_dyn}
    \begin{tabular}{l|l|l|l|l}
    \toprule
            & & H-OBCA & \makecell{TDR-\\OBCA} & OCEAN\\
        \hline \multirow{3}{*}{\begin{tabular}{l} 
        Steering \\
        Angle \\
        (rad)
        \end{tabular}} & Mean & -0.3382 & -0.3382 & -0.3487\\
        \cline { 2 - 5 } & Max & 0.1753 & 0.1753 & 0.0292 \\
        \cline { 2 - 5 } & Min & -0.5105 & -0.5105 & -0.5105\\
        \cline { 2 - 5 } & Std Dev & 0.2102 & 0.2102 & $\mathbf{0.1734}$\\
        \hline \hline \multirow{4}{*}{\begin{tabular}{l} 
        Velocity \\
        $\left(m / s\right)$
        \end{tabular}} & Mean & -0.8556 & -0.8556 & -0.6383\\
        \cline { 2 - 5 } & Max & 0.0000 & 0.0000 & -0.0300\\
        \cline { 2 - 5 } & Min & -1.0000 & -1.0000 & -0.7466\\
        \cline { 2 - 5 } & Std Dev & 0.2361 & 0.2361 & $\mathbf{0.1382}$\\
        \hline \hline \multirow{3}{*}{\begin{tabular}{l} 
        Acceleration \\
        $\left(m / s^2\right)$
        \end{tabular}} & Mean & 0.0095 & 0.0095 & -0.0529\\
        \cline { 2 - 5 } & Max & 0.3611 & 0.3611 & 0.2585\\
        \cline { 2 - 5 } & Min & -0.4057 & -0.4057 & -0.4000\\
        \cline { 2 - 5 } & Std Dev & 0.1265 & 0.1265 & $\mathbf{0.0851}$\\
        \hline
    \end{tabular}
\end{table}

In addition to robustness, we also compare the control outputs of the algorithms.
Fig.\ref{fig:apollo_vertical_dyn} shows the outputs of different algorithms in a typical scenario out of 80 cases.
From the trajectory plot and the corresponding kinematic curve plot, it can be seen that OCEAN generates trajectory that are almost identical to those obtained by H-OBCA or TDR-OBCA.

As a supplement to Fig.\ref{fig:apollo_vertical_dyn}, We compare the minimum, maximum and mean control outputs in Table.\ref{tab:compare_dyn}. 
On average, OCEAN generates trajectory with smaller variances in acceleration, velocity, and steering angle in comparison with TDR-OBCA, which indicates that the generated trajectory has better smoothness property.


\subsection{OCEAN Simulation results}\label{simulation_3}
To simulate OCEAN in different parking scenarios, we select parallel, vertical, and oblique parking spaces extracted from natural scenes for testing. Other vehicles are added as obstacles in the scenes. There are 116 parallel parking cases with 6-7 potential obstacles, 139 perpendicular parking cases with 8-10 potential obstacles, and 105 oblique parking cases. The position and orientation of other vehicles and ego vehicle are chosen randomly in each test cases. Illustrations of some of these cases are shown in Fig.\ref{fig:our_sim_test}.

\begin{table}
    \centering
    \captionsetup[table]{skip=20pt} 
    \caption{Trajectory evaluation results over the successful cases.}
    \label{tab:our_single_case}
    \begin{tabular}{c|c|c|c|c|c}
        \hline 
        Case ID & (a) & (b) & (c) & (d) & (e) \\
        \hline 
        $N_{OA}$ & 3 & 4 & 6 & 10 & 14 \\
        \hline 
        $N_{OB}$ & 3 & 3 & 5 & 8 & 10 \\
        \hline 
        $t_{f, TDR} (\mathrm{ms})$ & 1027.07 & 1395.43 & 1459.12 & 1691.88 & 1897.96 \\
        \hline 
        $t_{f, OCEAN} (\mathrm{ms})$ & 164.42 & 159.84 & 283.44 & 319.24 & 325.30 \\
        \hline 
        Improved Rate & $\mathbf{87.61\%}$ & $\mathbf{85.41\%}$ & $\mathbf{84.75\%}$ & $\mathbf{64.21\%}$ & $\mathbf{78.28\%}$ \\
        \hline
    \end{tabular}
\end{table}

\begin{table}
    \centering
    \captionsetup[table]{skip=20pt} 
    \caption{Calculation Times of Partial Sub-Problems in OCEAN. $obs\_var$ denotes solving the dual variable collision avoidance problem in \eqref{admm_1}; $speed\_pro$ denotes solving the speed and acceleration problem in \eqref{admm_2}; $time\_interval$ denotes solving the optimal time interval problem in \eqref{admm_3}; and $path\_pro$ denotes solving the MPC path plan problem in \eqref{admm_4}.}
    \label{tab:our_step_time}
    \begin{tabular}{c|c|c|c|c}
        \hline 
         & obs var & speed pro & time interval & path pro \\
        \hline 
        time(ms) & 104.66 & 1.3 & 0.74 & 7.37 \\
        \hline 

    \end{tabular}
\end{table}

We also study how solving time varies with the number of obstacles. $N_{OA}$ denotes the number of the vehicles and obstacles which are non-penetrable boundaries, while $N_{OB}$ denotes the number of additionally added virtual obstacles and virtual boundaries that are used to limit the trajectory generation area when the optimization space becomes too large. Table.\ref{tab:our_single_case} lists the solving time for H-OBCA, TDR-OBCA and OCEAN in different scenarios, as well as the number of obstacles and boundaries. 
In the five test scenarios, OCEAN achieves at least 60\% solution time reduction.

\begin{figure}
	\begin{center}
		\includegraphics[scale=0.6]{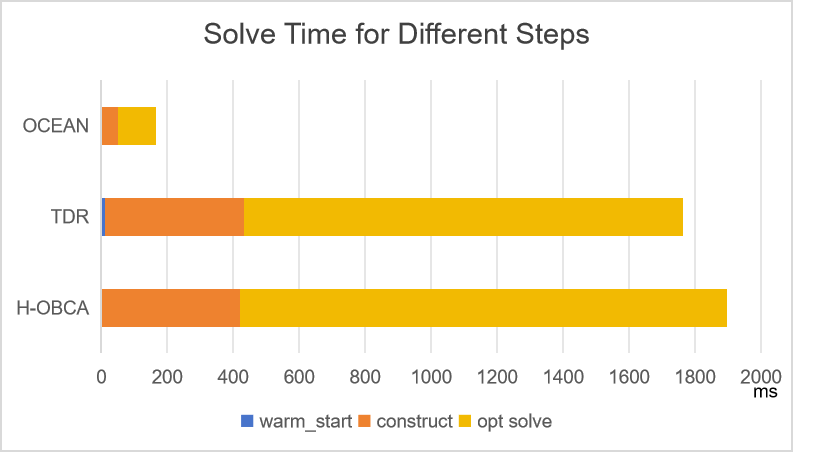}
	\end{center}
	\vspace{0.1cm}
	\caption{
		\label{fig:bev_view}
		Comparison of Time Consumption in Different Solution Steps for Three Algorithms.
	}
\end{figure}

Over 360 test cases are conducted and the results are shown in Fig.\ref{fig:bev_view} by histogram. The planning problem consists of warm start, optimization problem construction, and solving optimization problem. As mentioned in \ref{parallel_computation}, Warm-start is not necessary for OCEAN, while TDR-OBCA and H-OBCA highly depend on the trajectory quality generated by warm start process. 

On the other hand, it is worth noting that  H-OBCA and TDR-OBCA require additional time to formulate the NLP problem due to the use of the Casadi and Ipopt solvers, which consists of calculating the gradients and hessian of Lagrangian function and constraints. In contrast, OCEAN poses the ADMM sub-problems either as a SOCP or a QP problem, which can be solved by ECOS and OSQP. The problem formulation time can almost be negligible.

Table.\ref{tab:our_step_time} lists the detailed solving time for each sub-problem. 
In the table, it can be seen that the main time consumption comes from updating dual variables \eqref{admm_1} and solving MPC problem \eqref{admm_4}, which respectively take up approximately 63.63\% and 4.48\% of the total solving time. Thanks to ADMM problem decomposition and parallel optimization, our method takes averagely 160 ms to finish a planning loop, while TDR-OBCA and H-OBCA take more than 1200 ms.

\subsection{Real world road tests}\label{simulation_4}
We have performed hundreds of hours of driving tests on our auto-pilot platform to validate its robustness and efficiency. The overall upstream architecture for trajectory provider is similar to that described in \cite{tdr}. However, in the perception module, we only use cameras as an input source, following a purely visual approach.

\begin{figure}
	\begin{center}
		\includegraphics[scale=0.22]{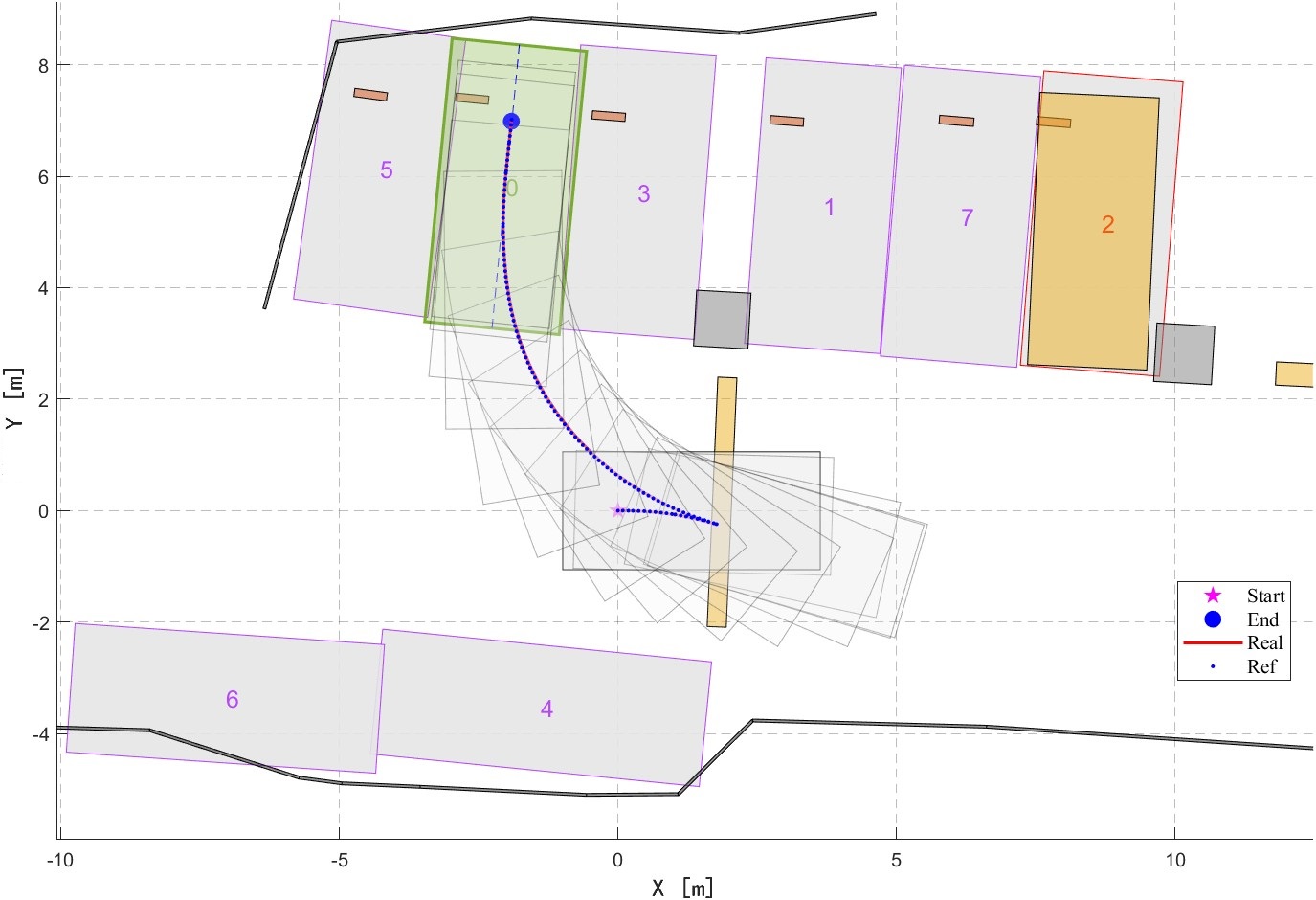}
	\end{center}
	\caption{
		\label{fig:real_vertical}
		Real Test Tracking Trajectory Visualization.
	}
\end{figure}

The three types of simulated scenarios presented in \ref{simulation_3} are selected from hundreds of hours of real road tests. Fig.\ref{fig:real_vertical}, for example, shows a typical vertical parking test scenario. For all road tests, the lateral control accuracy is high, ranging from 0.01 to 0.2 meters.

\begin{table}[ht]
\centering
\caption{Valet parking: Trajectory end pose accuracy on real road tests.}
\label{tab:real_test_error}
\begin{tabular}{l|ll}
\hline 
\text{Parking Scenarios (Partitions)} & \text{Lateral Error (m)} & \text{Heading Error (deg)} \\
\hline \hline 
\text{Vertical parking} & 0.0020 & 0.298 \\
\text{Parallel parking} & 0.0014 & 0.037 \\
\text{Oblique parking} & 0.0019 & 0.043 \\
\hline
\end{tabular}
\end{table}

Table.\ref{tab:real_test_error} shows the control performance of real world road tests, including the lateral error and the heading error at each trajectory endpoint. The controller of the autonomous vehicle follows the trajectory generated by OCEAN. The results confirm that the trajectories generated by OCEAN are smooth, subject to vehicle kinematic constraints, and result in lower tracking errors in real world road tests.

\section{CONCLUSIONS}\label{sec:conclusion}

In this paper, we propose and validate an autonomous parking trajectory optimization method called OCEAN, which demonstrates high efficiency and robustness in solving parking planning problems. Through validation on hundreds of simulation scenarios and real world parking lots, we have confirmed that this method outperforms other benchmark methods in terms of system performance. Our method makes it possible to deploy large-scale parking planner on low-computing power platforms that require real-time performance.

\addtolength{\textheight}{-12cm}   










\begin{thebibliography}{99}

\bibitem{curve_parking} Liang, Zhao, Guoqiang Zheng, and Jishun Li. "Automatic parking path optimization based on bezier curve fitting." 2012 IEEE International Conference on Automation and Logistics. IEEE, 2012.
\bibitem{point_mass_collision_avoidance}Alonso-Mora, Javier, et al. "Collision avoidance for aerial vehicles in multi-agent scenarios." Autonomous Robots 39 (2015): 101-121.
\bibitem{MIP} da Silva Arantes, Marcio, et al. "Collision-free encoding for chance-constrained nonconvex path planning." IEEE Transactions on Robotics 35.2 (2019): 433-448.
\bibitem{linear_prog}Wang, Qianhao, et al. "A linear and exact algorithm for whole-body collision evaluation via scale optimization." 2023 IEEE International Conference on Robotics and Automation (ICRA). IEEE, 2023.
\bibitem{esdf}Oleynikova, Helen, et al. "Signed distance fields: A natural representation for both mapping and planning." RSS 2016 workshop: geometry and beyond-representations, physics, and scene understanding for robotics. University of Michigan, 2016.
\bibitem{h_obca}Zhang, Xiaojing, et al. "Autonomous parking using optimization-based collision avoidance." 2018 IEEE Conference on Decision and Control (CDC). IEEE, 2018.
\bibitem{obca}Zhang, Xiaojing, Alexander Liniger, and Francesco Borrelli. "Optimization-based collision avoidance." IEEE Transactions on Control Systems Technology 29.3 (2020): 972-983.
\bibitem{tdr} He, Runxin, et al. "TDR-OBCA: A reliable planner for autonomous driving in free-space environment." 2021 American Control Conference (ACC). IEEE, 2021.
\bibitem{rda} Han, Ruihua, et al. "Rda: An accelerated collision free motion planner for autonomous navigation in cluttered environments." IEEE Robotics and Automation Letters 8.3 (2023): 1715-1722.
\bibitem{mpc}Erlien, Stephen M., Susumu Fujita, and J. Christian Gerdes. "Safe driving envelopes for shared control of ground vehicles." IFAC Proceedings Volumes 46.21 (2013): 831-836.
\bibitem{spheres}Chen, Jianyu, Wei Zhan, and Masayoshi Tomizuka. "Constrained iterative lqr for on-road autonomous driving motion planning." 2017 IEEE 20th International conference on intelligent transportation systems (ITSC). IEEE, 2017.
\bibitem{corridor}Liu, Sikang, et al. "Planning dynamically feasible trajectories for quadrotors using safe flight corridors in 3-d complex environments." IEEE Robotics and Automation Letters 2.3 (2017): 1688-1695.
\bibitem{corridor2}Jiang, Yuncheng, et al. "Robust Online Path Planning for Autonomous Vehicle Using Sequential Quadratic Programming." 2022 IEEE Intelligent Vehicles Symposium (IV). IEEE, 2022.
\bibitem{double sd}Schulman, John, et al. "Motion planning with sequential convex optimization and convex collision checking." The International Journal of Robotics Research 33.9 (2014): 1251-1270.
\bibitem{risk field}Tan, Shuhang, Zhiling Wang, and Yan Zhong. "RCP-RF: A Comprehensive Road-car-pedestrian Risk Management Framework based on Driving Risk Potential Field." arXiv preprint arXiv:2305.02493 (2023).
\bibitem{sdf any}Zhang, Tingrui, et al. "Continuous Implicit SDF Based Any-shape Robot Trajectory Optimization." arXiv preprint arXiv:2303.01330 (2023).
\bibitem{edge computing}Li, Guoliang, et al. "Edge Accelerated Robot Navigation with Hierarchical Motion Planning." *arXiv preprint arXiv:2311.08983* (2023).
\bibitem{admm}S. Boyd, N. Parikh, E. Chu, B. Peleato, J. Eckstein, *et al.*, “Distributed optimization and statistical learning via the alternating direction method of multipliers,” *Found. Trends Mach. Learn.*, vol. 3, no. 1,pp. 1–122, 2011.
\bibitem{admm ilqr} Huang, Zhenmin, Shaojie Shen, and Jun Ma. "Decentralized iLQR for Cooperative Trajectory Planning of Connected Autonomous Vehicles via Dual Consensus ADMM." *arXiv preprint arXiv:2301.04386* (2023).
\bibitem{ipm}  Domahidi, Alexander, Eric Chu, and Stephen Boyd. "ECOS: An SOCP solver for embedded systems." 2013 European control conference (ECC). IEEE, 2013.
\bibitem{app_socp} Lobo, M. S., Vandenberghe, L., Boyd, S., \& Lebret, H. (1998). Applications of second-order cone programming. Linear Algebra and Its Applications, 193–228. https://doi.org/10.1016/s0024-3795(98)10032-0
\bibitem{mars landing} Acikmese, B., \& Ploen, S. R. (2007). Convex Programming Approach to Powered Descent Guidance for Mars Landing. Journal of Guidance, Control, and Dynamics, 1353–1366. https://doi.org/10.2514/1.27553
\bibitem{iaps} Zhou, Jinyun, et al. "Dl-iaps and pjso: A path/speed decoupled trajectory optimization and its application in autonomous driving." arXiv preprint arXiv:2009.11135 (2020).
\bibitem{rs} Fraichard, Thierry, and Alexis Scheuer. "From Reeds and Shepp's to continuous-curvature paths." IEEE Transactions on Robotics 20.6 (2004): 1025-1035.

\end{thebibliography}
\end{document}